\documentclass{article}

\usepackage{microtype}
\usepackage{graphicx}
\usepackage{subcaption}
\usepackage{booktabs} 
\usepackage{pifont}
\usepackage{xcolor}
\usepackage{xspace}
\usepackage{enumitem}
\usepackage{bbm}

\usepackage{hyperref}

\definecolor{ourcolor}{HTML}{8B4513}
\newcommand{\ours}{\textcolor{ourcolor}{\textsc{Opcd}}\xspace}

\usepackage[accepted]{icml2026}

\usepackage[utf8]{inputenc} 
\usepackage[T1]{fontenc}    
\usepackage{hyperref}       
\usepackage{url}            
\usepackage{booktabs}       
\usepackage{amsfonts}       
\usepackage{nicefrac}       
\usepackage{microtype}      
\usepackage{xcolor}         
\usepackage{amssymb}
\usepackage{graphicx}

\usepackage{tabularx}      
\usepackage[table]{xcolor}  
\usepackage{amsmath}
\usepackage{amssymb}
\usepackage{mathtools}
\usepackage{amsthm}

\usepackage{booktabs}
\usepackage{multirow}
\usepackage{makecell}
\usepackage[table]{xcolor}
\usepackage{graphicx}
\usepackage{algorithm}
\usepackage{algorithmic}
\usepackage{amsmath}
\usepackage{bbm}

\usepackage[capitalize,noabbrev]{cleveref}

\theoremstyle{plain}

\theoremstyle{definition}

\theoremstyle{remark}

\usepackage[textsize=tiny]{todonotes}

\icmltitlerunning{Weak Critics Make Strong Learners: On-Policy Critique Distillation for Scalable Oversight}

\begin{document}

\twocolumn[
  \icmltitle{Weak Critics Make Strong Learners: \\ On-Policy Critique Distillation for Scalable Oversight}



  \icmlsetsymbol{equal}{*}
  \icmlsetsymbol{equal_advise}{\dag}

  \begin{icmlauthorlist}
    \icmlauthor{Can Jin}{equal,rutgers}
    \icmlauthor{Jiakang Li}{equal,rutgers}
    \icmlauthor{Rui Wu}{rutgers}
    \icmlauthor{Eddy Z. Zhang}{equal_advise,rutgers}
    \icmlauthor{Dimitris N. Metaxas}{equal_advise,rutgers}
  \end{icmlauthorlist}

  \icmlaffiliation{rutgers}{Department of Computer Science, Rutgers University, Piscataway, United States}

\icmlcorrespondingauthor{Can Jin}{can.jin@rutgers.edu}
\icmlcorrespondingauthor{Jiakang Li}{jiakang.li@rutgers.edu}

  \icmlkeywords{Scalable Oversight, Weak-to-Strong Generalization, LLM Alignment, LLM Reasoning}

  \vskip 0.3in
]



\printAffiliationsAndNotice{\icmlEqualContribution, $^{\dag}$ Equal Advising.}  

\begin{abstract}
As large language models become stronger, weak supervisors may fail to provide reliable labels, preferences, or final judgments for complex outputs, limiting both weak-to-strong generalization and scalable oversight. We study a more tractable form of weak supervision: using a weak model as a critic rather than as a labeler or judge. Instead of solving the task or selecting the correct answer, the weak critic only needs to provide a non-misleading revision direction that helps the strong model better use its own knowledge. We call this setting \emph{weak-critic strong oversight}. We first show that weak critiques can improve frozen strong models at inference time, and that critique quality is key to this improvement. We then propose progressive on-policy critique distillation (\ours), which filters high-quality critiques and distills critic-guided behavior into the strong model through adaptive self-teacher signals. Experiments on reasoning and alignment benchmarks show that our method improves strong models over training epochs, suggesting an effective path for scalable oversight with weak supervision. Code is available at: \textcolor{blue}{\url{https://github.com/jiakanglee/OPCD-Weak_critiques_strong}}
\end{abstract}

\section{Introduction}
Modern large language models (LLMs) are commonly aligned with human supervision, such as task demonstrations, preference labels, reward models, and reinforcement learning from human feedback~\citep{christiano2017deep,ouyang2022training, bai2022training}. These methods work well when the supervisor can reliably judge the model output. However, as models become stronger, they may produce answers, plans, proofs, or code that are difficult for humans or weaker models to fully verify. This creates a central challenge for alignment: how can a weak supervisor guide a stronger model when the final task is too hard for the supervisor to solve or judge?

Two related research directions study this challenge. Weak-to-strong generalization asks whether supervision from a weak model can elicit useful behavior from a stronger pretrained model~\citep{burns2023weak}. Scalable oversight asks how a weak human or model can provide reliable supervision for a stronger system, often through assistance, interaction, or debate~\citep{amodei2016concrete,irving2018debate,bowman2022measuring,khan2024debating}. Although these directions differ in their goals, many existing methods place a similar burden on the weak supervisor: the weak model must provide direct labels, soft logits, preference signals, or final judgments over complete answers. This can be too demanding when the task is beyond the weak supervisor's full ability. In such cases, the supervision signal can be noisy, incomplete, or systematically wrong, which limits both weak-to-strong generalization and scalable oversight.

In this paper, we study a different form of weak supervision: using the weak model as a \emph{critic} rather than as a labeler or judge. A weak critic does not need to solve the task, provide the correct answer, identify every error, or give a detailed revision plan. It can be useful by giving a general but correct revision direction, such as suggesting that the reasoning is incomplete, a condition is missing, a boundary case should be checked, or the response should be safer. This form of feedback is often easier than labeling or judging complete answers, and is close to common human--AI interaction: users may not know the full solution, but they can still give feedback that helps a stronger model revise. As long as the critique is not misleading, it can help the strong model better use its own knowledge without requiring the weak supervisor to provide full supervision. We call this setting \emph{weak-critic strong oversight}.

We first test this idea at inference time. Given a question, the strong model produces an initial answer, the weak model critiques it, and the strong model then revises its answer conditioned on the question, the initial answer, and the critique. This directly evaluates whether weak critiques can improve a frozen strong model. Our results show that they can, even when the critique only gives a general revision direction rather than a detailed error analysis. This supports the core hypothesis of \emph{weak-critic strong oversight}: weak supervisors may not need to solve or judge the full task to provide useful oversight. We also find that critique quality is central. Helpful critiques improve performance, while misleading critiques can hurt performance, even compared with using no critique. This motivates filtering useful critiques before using them for training.

To internalize the inference-time improvement, we propose a progressive \textbf{o}n-\textbf{p}olicy \textbf{c}ritique \textbf{d}istillation method (\ours). In each epoch, the current strong model generates on-policy answers, and the weak model critiques these answers. We then use an outcome- and rubric-based quality metric to keep only useful critiques. For each kept example, the critic-conditioned strong model serves as a self-teacher, using the critique as guidance to provide dense token-level signals. The student is the same strong model without access to the critique, trained by on-policy distillation. After each update, the strong model produces new answers with new error patterns, and the weak critic provides fresh critiques for the updated model. This process distills the useful critic-guided behavior observed at inference time while keeping the supervision adaptive to the current strong model.

Our experiments show that \emph{weak-critic strong oversight} improves strong-model performance in both inference-time and training-time settings. Compared with standard weak-to-strong methods that directly distill weak-model responses or logits, our method does not force the weak model to provide full supervision. Compared with ground-truth supervised finetuning, our method studies a more realistic oversight setting where reliable labels may be unavailable or too hard for weak supervisors to provide. Across reasoning and alignment benchmarks, progressive on-policy critique distillation improves the strong model over training epochs, showing that critique-based supervision is an effective path for scalable oversight and weak-to-strong generalization.

Our main contributions are:
\begin{itemize}[leftmargin=1.5em]
\item[\ding{72}] \textbf{Critique-based weak supervision.} We identify critiquing as a more tractable form of weak supervision than labeling or judging, and propose \emph{weak-critic strong oversight} for scalable oversight and weak-to-strong generalization.

\item[\ding{72}] \textbf{Inference-time validation.} We show that weak critiques can improve strong-model performance at inference time, even when they provide only general revision directions, and find that critique quality is key to reliable improvement.

\item[\ding{72}] \textbf{Progressive critique distillation.} We introduce \ours, a progressive on-policy critique distillation strategy that filters high-quality critiques and uses them as adaptive weak feedback to train the strong model.

\item[\ding{72}] \textbf{Strong  results.} Across multiple benchmarks, \ours{} progressively improves strong-model performance, showing that critique-based supervision can effectively use weak oversight for stronger models.

\end{itemize}

\section{Related Works}

\paragraph{Scalable oversight and weak-to-strong generalization.} Scalable oversight aims to develop methods to supervise AI systems on tasks that are difficult for humans \cite{amodei2016concrete, bowman2022measuring}. The primary focus has been on designing human-AI collaboration protocols that help humans evaluate AI outputs more accurately, for example through debate and consultancy \cite{irving2018debate, kenton2024scalable}, and on lowering the cognitive burden of evaluation through critique-assisted review and prover-verifier games \cite{saunders2022selfcritique, mcaleese2024llmcritic, kirchner2024prover}. In contrast, weak-to-strong generalization (W2S) \cite{burns2023weak} explores a complementary direction that designs learning algorithms which let a strong pretrained model generalize correctly from weak supervision as if it had been trained on higher-quality labels. A growing body of work strengthens this elicitation through iterative label refinement, easy-to-hard reward transfer, weak-LLM preference labeling, internal-coherence elicitation, and self-consistency filtering for reasoning \cite{ye2025iterative, sun2024easy, tao2024your, wen2025unsupervised, yang2024weakstrongreasoning, jin2024learning}, with inference-time variants contrasting weak and strong distributions or guiding decoding with weak step-level scores \cite{li2023contrastive, ding2025w2s}. Theoretical analyses bound the W2S gain by the strong model's misfit on weak labels and characterize pseudolabel correction under an expansion condition \cite{charikar2024quantifying, lang2024theoretical}, while \citet{yang2025superficial} document a failure mode in which strong models pass weak supervision on prompts the supervisor knows but remain misaligned where it does not. Our setting follows the W2S protocol of \citet{burns2023weak} but uses weak critiques in place of weak labels and targets generative reasoning rather than classification.

\paragraph{LLM Alignment.} Aligning LLMs with human preferences is most commonly done via reinforcement learning from human feedback \cite{christiano2017deep, ouyang2022training, bai2022training}, which trains a reward model on preference comparisons and optimizes the policy with PPO \cite{schulman2017ppo}; direct preference methods bypass the reward model and train on preferences end-to-end \cite{rafailov2023dpo}. Because human supervision is expensive and difficult to scale, a separate line replaces or supplements it with model-generated signals: AI feedback judged against written principles \cite{bai2022constitutional, lee2024rlaif, jin2026reasoning}, self-judgment loops in which the policy itself plays the role of judge \cite{yuan2024self, wu2024meta}, and dedicated critic models that surface mistakes the policy or human annotators might miss \cite{mcaleese2024llmcritic, ankner2024critique}. These methods take the supervisor's signal as a training target. We instead treat it as a critique and use only those critiques that demonstrably improve the policy's answer.

\paragraph{LLM Reasoning.} Eliciting reasoning from LLMs has been driven by chain-of-thought prompting and its inference-time variants such as self-consistency and tree-search decoding \cite{wei2022chain, wang2023selfconsistency, yao2023tree}. Self-improvement methods finetune the policy on its own successful traces \cite{zelikman2024star} or train process verifiers on step-level correctness to densify the reward on hard problems \cite{lightman2024verify, wang2024mathshepherd, jin2025your}. A complementary line teaches the model to revise its outputs, either at inference time through verbal feedback \cite{madaan2023selfrefine, shinn2023reflexion} or during training through multi-turn reinforcement learning \cite{kumar2024score, qu2024recursive}. Recent large-scale efforts show that pure outcome-based RL on a strong base model can elicit long chain-of-thought reasoning at scale \cite{deepseekai2025r1}, while online distillation has shifted toward student-side losses that match the teacher under the student's own distribution \cite{gu2023minillm,Sun_2026_CVPR}. These methods either require ground-truth verification or assume a critic at least as strong as the policy.

\section{Preliminary Inference-Time Investigation}
\label{sec:inference_time}

Whether a weak model can provide useful critiques to improve a stronger model at inference time is the foundation of \ours framework. In this section, we conduct a preliminary investigation to answer two questions: \\

(i) Can weak-model critique improve strong-model performance beyond sampling more responses, and does this effect generalize across reasoning and alignment tasks as well as across thinking and non-thinking models?\\
(ii) Does the quality of the critique affect the final accuracy?

\paragraph{Experimental Setting.}
We evaluate the critique-and-refine paradigm on both non-thinking and thinking models. For non-thinking models, we use Phi-4-mini-instruct-3.8B~\cite{abdin2024phi} as the weak model and Phi-4-14B~\cite{abdin2024phi} as the strong model. The experiments are conducted on GPQA Diamond~\cite{rein2023gpqa} for reasoning task and IFEval~\cite{zhou2023instruction} for instruction-following alignment task. For thinking models, we use Qwen3-1.7B~\cite{yang2025qwen3} as the weak model and Qwen3-8B~\cite{yang2025qwen3} as the strong model. Both models are evaluated with thinking mode enabled on AIME 2024~\cite{aime2024_i,aime2024_ii} and AIME 2025~\cite{aime2025_i,aime2025_ii}, containing 60 problems in total. This setting tests whether the usefulness of weak-model critiques also scales to thinking models.

For each problem, the critique-and-refine pipeline consists of three stages and we denote it as \textit{S+W critic+refine} for easier annotation. First, the strong model generates an initial answer. Then, the weak model receives the original problem and the strong model's initial answer, and produces a critique. Finally, the strong model refines its answer based on the problem, the initial answer, and the weak-model critique. This process is repeated for up to 16 independent chains as reported in Table~\ref{tab:inference_time_all}. Formally, for each chain $i$, the process is written as:
\begin{equation}
\label{eq:critique_refine}
a_i = S(x), \quad c_i = W(x,a_i), \quad r_i = S(x,a_i,c_i),
\end{equation}
where $a_i$, $c_i$, and $r_i$ denote the initial answer, weak-model critique, and refined answer of chain $i$, respectively. For an inference budget of $k \in \{1,2,4,8,16\}$ chains, we report pass@$k$, where the prediction is counted as correct if any of the first $k$ final answers is correct.

\begin{figure}[t]
    \centering
    \includegraphics[width=\linewidth]{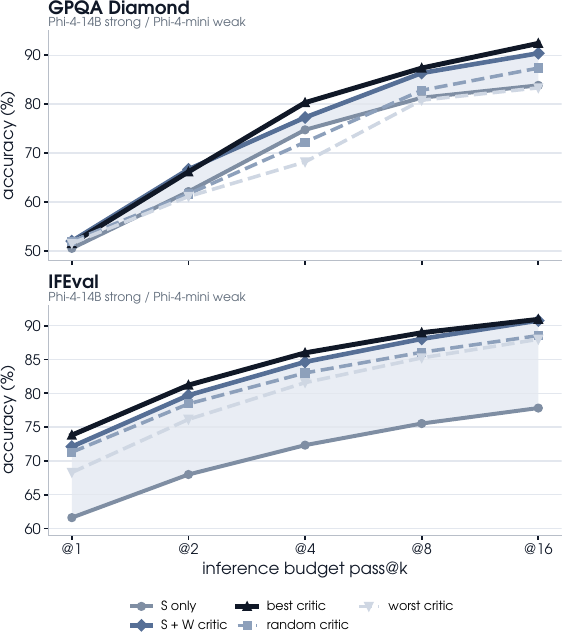}
    \caption{Pass@k performance reported as accuracy percentage under different inference budgets on GPQA Diamond and IFEval. The strong model is Phi-4-14B and the weak model is Phi-4-mini-instruct. The shaded region highlights the performance gap between the $S only$ baseline and the $S+W critic+refine$ setting.}
    \label{fig:Phi4_ifeval_GPQA}
\end{figure}

\begin{table*}[t]
\centering
\vspace{-0.4em}
\small
\renewcommand{\arraystretch}{1.12}
\setlength{\tabcolsep}{5.2pt}
\begin{tabular}{ll l ccccc}
\toprule
\multirow{2}{*}{Benchmark}
& \multirow{2}{*}{Model Pair}
& \multirow{2}{*}{Method}
& \multicolumn{5}{c}{Inference Budget} \\
\cmidrule(lr){4-8}
& & & pass@1 & pass@2 & pass@4 & pass@8 & pass@16 \\
\midrule

\multirow{5}{*}{GPQA Diamond}
& \multirow{5}{*}{\makecell[l]{Phi-4-14B (S)\\Phi-4-mini-instruct (W)}}
& \textit{$S$ only}
& 50.51 & 62.12 & 74.75 & 81.31 & 83.84 \\
& & \textit{$S$+$W$ critic+refine}
& \textbf{51.99} & \textbf{66.67} & 77.27 & 86.36 & 90.40 \\
& & \textit{$S$+best critic+refine}
& 51.52 & 66.16 & \textbf{80.30} & \textbf{87.37} & \textbf{92.42} \\
& & \textit{$S$+random critic+refine}
& \textbf{51.99} & 61.62 & 72.22 & 82.83 & 87.37 \\
& & \textit{$S$+worst critic+refine}
& 51.52 & 61.11 & 68.18 & 80.81 & 83.33 \\

\midrule

\multirow{5}{*}{IFEval}
& \multirow{5}{*}{\makecell[l]{Phi-4-14B (S)\\Phi-4-mini-instruct (W)}}
& \textit{$S$ only}
& 61.61 & 67.99 & 72.33 & 75.53 & 77.82 \\
& & \textit{$S$+$W$ critic+refine}
& 72.11 & 79.71 & 84.64 & 88.05 & 90.76 \\
& & \textit{$S$+best critic+refine}
& \textbf{73.81} & \textbf{81.21} & \textbf{85.98} & \textbf{88.96} & \textbf{90.94} \\
& & \textit{$S$+random critic+refine}
& 71.29 & 78.44 & 83.01 & 86.07 & 88.54 \\
& & \textit{$S$+worst critic+refine}
& 68.29 & 76.15 & 81.58 & 85.24 & 87.99 \\

\midrule

\multirow{3}{*}{AIME24+25}
& \multirow{3}{*}{\makecell[l]{Qwen3-8B (S)\\Qwen3-1.7B (W)}}
& \textit{$S$ only}
& 71.67 & 81.67 & 81.67 & 83.33 & 86.67 \\
& & \textit{$W$ only}
& 40.00 & 51.67 & 61.67 & 66.67 & 73.33 \\
& & \textit{$S$+$W$ critic+refine}
& \textbf{75.00} & \textbf{81.67} & \textbf{81.67} & \textbf{85.00} & \textbf{90.00} \\

\bottomrule
\end{tabular}
\caption{Inference-time scaling results on reasoning and alignment benchmarks.
All results are reported as accuracy percentages. $S$ and $W$ denote the strong and weak models, respectively. Best results within each benchmark are highlighted in bold.}
\label{tab:inference_time_all}
\vspace{-0.4em}
\end{table*}

\paragraph{Baselines.}
We compare the proposed critique-and-refine pipeline with several inference-time baselines. The basic baseline is \textit{S only}, where the strong model independently generates multiple answers without receiving any critique.

For the non-thinking model Phi-4 experiments, we further introduce three baselines to analyze how critique quality affects final performance. To study the critic quality effect alone, these baselines are constructed from cached outputs for the complete $k$ chains in \textit{S+ W critic + refine}. For each inference budget $k$, we select one critique from the first $k$ chains and pair it with each of the $k$ initial answers before refinement. In \textit{S + best critic + refine}, we select a critique from a chain where the initial answer is incorrect but the refined answer becomes correct, if such a chain exists. This setting approximates the effect of a high-quality critique. In \textit{S + random critic + refine}, we uniformly sample one critique from the first $k$ chains, which measures the effect of critique reuse without correctness information. In \textit{S + worst critic + refine}, we select a critique from a chain where the initial answer is correct but the refined answer becomes incorrect, if such a chain exists. For the thinking-model Qwen experiments, we also report \textit{W only} where the weak model independently generates multiple answers as \textit{S only}. We do not include other baselines as the Phi-4 experiments in the thinking model setting, because this experiment is intended as a focused test of whether weak critique can scale thinking-mode generation rather than a complete critique-quality analysis.

\paragraph{Weak Model Critique Framework Improves Strong Model Performance Across Tasks and Model Types.}
As shown in Table~\ref{tab:inference_time_all} and Figure~\ref{fig:Phi4_ifeval_GPQA}, weak-model critique improves strong-model performance at inference time across both reasoning and alignment tasks. On GPQA Diamond, \textit{S only} achieves 50.51 pass@1 and 83.84 pass@16, while \textit{S+W critic+refine} improves the results to 51.99 pass@1 and 90.40 pass@16. Figure~\ref{fig:Phi4_ifeval_GPQA} further shows that this improvement is maintained as the inference budget increases, indicating that weak critique provides useful feedback signals beyond sampling more responses. A similar trend appears on IFEval. As reported in Table~\ref{tab:inference_time_all}, the \textit{S only} baseline achieves 61.61 pass@1 and 77.82 pass@16. With weak-model critique and refinement, the performance increases to 72.11 pass@1 and 90.76 pass@16. The scaling trend in Figure~\ref{fig:Phi4_ifeval_GPQA} also shows a consistent gap between \textit{S only} and \textit{S+W critic+refine} across different inference budgets where the strong model can consistently outperform \textit{S only}. These observations suggest that weak model critique framework improves strong model performance across reasoning and alignment tasks.

We further examine whether weak-model critique remains useful under thinking-mode generation. As shown in Table~\ref{tab:inference_time_all}, on AIME 2024 and AIME 2025, the weak model itself is substantially weaker than the strong model, achieving 40.00 pass@1 compared with 71.67 pass@1 for the strong model. Nevertheless, when used as a critic, the weak model improves the strong model from 71.67 to 75.00 at pass@1 and from 86.67 to 90.00 at pass@16. These results suggest that weak model critique improves strong model performance across thinking and non-thinking models.

\paragraph{High-Quality Critiques Improve Performance and Generalize Across different Initial Answers.}
The baselines for Phi-4 in Table~\ref{tab:inference_time_all} show that critique quality matters in improving performance. Notably, on GPQA Diamond \textit{$S$+best critic+refine} achieves best performance on pass@4, pass@8, and pass@16 with 80.30, 87.37 and 92.42 respectively. While we can also observe that \textit{$S$+random critic+refine} is consistently weaker than \textit{$S$+$W$ critic+refine} at larger budgets and \textit{$S$+worst critic+refine} performs the worst among other critique-quality baselines, with 68.18 pass@4, 80.81 pass@8, and 83.33 pass@16, eventually falling slightly below \textit{$S$ only} at pass@16. Also, a similar pattern appears on IFEval, \textit{$S$+best critic+refine} achieves the strongest results across all inference budgets, reaching 73.81 pass@1, 85.98 pass@4, and 90.94 pass@16, and is followed by \textit{$S$+random critic+refine} and \textit{$S$+worst critic+refine}. These results indicate that quality of critique matters, high-quality critique can better improve the performance in the framework but low-quality critique can weaken the effect of improvement across different tasks.

Since the selected critiques in baselines are not necessarily paired with the current initial answer, this result also suggests that a high-quality critique can generalize across different initial answers. In other words, useful critiques may provide transferable guidance, such as identifying common reasoning errors, highlighting missing constraints, or suggesting a more reliable solving direction, rather than only correcting one specific response.

\section{Method}
\subsection{Problem Definition and Notation}
\label{sec:problem_definition}

\begin{figure*}[t]
    \centering
    \includegraphics[width=\textwidth]{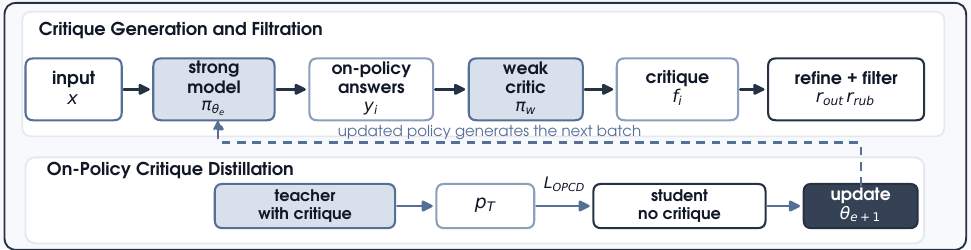}
    \caption{Overview of the \ours pipeline.}
    \label{fig:pipeline}
\end{figure*}

The overall framework is shown in Figure~\ref{fig:pipeline}. We consider a weak-to-strong oversight setting with a weak model $\pi_{\mathrm{w}}$ and a strong model $\pi_{\theta}$. Given an input question or instruction $x \in \mathcal{X}$, our goal is to improve the strong model using weak supervision, without requiring the weak model to provide the correct answer, a preference label, or a final judgment. Instead, the weak model acts as a critic that gives feedback on the strong model's own answers.

At training epoch $e$, let $\pi_{\theta_e}$ denote the current strong model before the update. For each input $x$, we sample a group of $G$ on-policy answers
\[
    y_i \sim \pi_{\theta_e}(\cdot \mid x),
    \qquad i=1,\ldots,G,
\]
where $y_i=(y_{i,1},\ldots,y_{i,T_i})$ and $y_{i,<t}$ denotes the prefix before token $y_{i,t}$. The weak model then critiques each answer:
\[
    f_i \sim \pi_{\mathrm{w}}(\cdot \mid x, y_i).
\]
Each critique $f_i$ provides a revision direction for the corresponding answer $y_i$. It does not need to give a full solution or a detailed correction; it only needs to offer useful and non-misleading guidance that can help the strong model better use its own knowledge.

Our key idea is to use the critic-conditioned strong model as a self-teacher. For each pair $(y_i,f_i)$, the teacher is the frozen current strong model conditioned on the weak critique,
\[
    \pi_{\theta}(\cdot \mid x, f_i, y_{i,<t}),
\]
while the student is the trainable strong model without access to the critique,
\[
    \pi_{\theta}(\cdot \mid x, y_{i,<t}).
\]
The training objective distills the critic-guided behavior of the teacher into the student, so that the strong model can internalize the benefit of weak critiques and improve without requiring critiques at test time.

\subsection{Critique Generation and Filtration}

Given the on-policy answers $\{y_i\}_{i=1}^{G}$ from the current strong model $\pi_{\theta_e}$, the weak critic produces a critique $f_i$ for each answer. Since weak critiques can be helpful or misleading, we filter them before using them for training. The goal of filtration is to keep only critiques that provide useful guidance for the strong model.

For each triple $(x,y_i,f_i)$, we first ask the current strong model to generate a refined answer conditioned on the question, the original answer, and the critique:
\[
    \hat{y}_i \sim \pi_{\theta_e}(\cdot \mid x, y_i, f_i).
\]
We then apply an outcome-based check to test whether the refined answer is correct:
\[
    r_{\mathrm{out}}(x,\hat{y}_i)
    =
    \mathbbm{1}\{\hat{y}_i \text{ is correct for } x\}.
\]
This check directly measures whether the critique helps the strong model revise toward a correct answer. However, outcome correctness alone may keep critiques that are vague, irrelevant, or not actually responsible for the improvement. Therefore, we also use a rubric-based check:
\[
\begin{aligned}
    r_{\mathrm{rub}}(x,y_i,f_i)
    &=
    \mathbbm{1}\Big\{
    \substack{
    f_i \text{ is relevant and useful as revision guidance}
    }
    \Big\}.
\end{aligned}
\]
A critique is kept only if it passes both checks:
\[
    h_i
    =
    r_{\mathrm{out}}(x,\hat{y}_i)
    \cdot
    r_{\mathrm{rub}}(x,y_i,f_i).
\]
The filtered training set for epoch $e$ is then
\[
    \mathcal{S}_e
    =
    \{(x,y_i,f_i): h_i = 1\}.
\]
Thus, the strong model is trained only on questions and on-policy answers with high-quality critiques. This filtration step is important because \ours{} aims to distill useful critic-guided behavior, rather than imitate weak feedback blindly.

\subsection{Progressive On-Policy Critique Distillation}
After obtaining the filtered set $\mathcal{S}_e$, we distill the critic-guided behavior into the strong model. For each selected triple $(x,y,f)$, we use the critic-conditioned strong model defined in Subsection \ref{sec:problem_definition} as the self-teacher, and train the same strong model without access to the critique as the student. Both are evaluated along the same on-policy answer trajectory $y$, so the critique only affects the teacher distribution while the student learns to internalize this guidance.

We optimize the student by minimizing the token-level KL divergence:
\[
\begin{aligned}
    \mathcal{L}_{\mathrm{\ours}}(\theta)
    &=
    \frac{1}{|\mathcal{S}_e|}
    \sum_{(x,y,f)\in \mathcal{S}_e}
    \sum_{t=1}^{|y|}
    \mathrm{KL}\!\Big(
        \pi_{\theta}(\cdot \mid x, y_{<t})
        \,\Big\| \\
    &\qquad\qquad
        \mathrm{stopgrad}\!\left[
        \pi_{\theta}(\cdot \mid x, f, y_{<t})
        \right]
    \Big).
\end{aligned}
\]
where the KL divergence is defined as
\[
    \mathrm{KL}(p \| q)
    =
    \sum_{v \in \mathcal{V}_{\mathrm{KL}}}
    p(v)\log\frac{p(v)}{q(v)},
\]
with $\mathcal{V}_{\mathrm{KL}}$ denoting the token set used for the KL computation, which can be the full vocabulary or a selected subset of tokens. This objective transfers the dense token-level guidance induced by the weak critique into the strong model, so that the updated model can benefit from critic-guided reasoning without requiring critiques at test time.

The training process is progressive and on-policy. Before each epoch, the current strong model generates new answers, the weak model critiques these answers, and the filtration step constructs a new high-quality set $\mathcal{S}_e$. The model is then updated with $\mathcal{L}_{\mathrm{\ours}}$. After the update, the next model $\pi_{\theta_{e+1}}$ produces new answers with different error patterns, leading to new critiques and a new filtered training set. In this way, the policy, critiques, and training data are updated together throughout training.  This makes \ours{} adaptive to the current strong model rather than relying on a fixed offline set of weak critiques. We summarize the full procedure in Algorithm \ref{alg:pocd}.

\begin{algorithm}[t]
\caption{\ours}
\label{alg:pocd}
\begin{algorithmic}[1]
\REQUIRE Strong model $\pi_{\theta_0}$, weak critic $\pi_{\mathrm{w}}$, training set $\mathcal{X}$, epochs $E$, group size $G$
\STATE $\mathcal{S}_{-1} \leftarrow \emptyset$
\FOR{epoch $e = 0, 1, \ldots, E-1$}
    \STATE $\mathcal{S}_e \leftarrow \mathcal{S}_{e-1}$
    \FOR{each input $x \in \mathcal{X}$}
        \STATE Sample on-policy answers $\{y_i \sim \pi_{\theta_e}(\cdot \mid x)\}_{i=1}^{G}$
        \STATE Generate weak critiques $\{f_i \sim \pi_{\mathrm{w}}(\cdot \mid x, y_i)\}_{i=1}^{G}$
        \FOR{$i = 1, \ldots, G$}
            \STATE Refine: $\hat{y}_i \sim \pi_{\theta_e}(\cdot \mid x, y_i, f_i)$
            \IF{$r_{\mathrm{out}}(x, \hat{y}_i) \cdot r_{\mathrm{rub}}(x, y_i, f_i) = 1$}
                \STATE $\mathcal{S}_e \leftarrow \mathcal{S}_e \cup \{(x, y_i, f_i)\}$
            \ENDIF
        \ENDFOR
    \ENDFOR
    \STATE Update $\theta_e \to \theta_{e+1}$ by minimizing $\mathcal{L}_{\mathrm{\ours}}(\theta)$ on $\mathcal{S}_e$
\ENDFOR
\STATE \textbf{return} $\pi_{\theta_E}$
\end{algorithmic}
\end{algorithm}

\section{\ours Training-Time Experiments}

We evaluate our method in two weak-to-strong scenarios: reasoning and alignment. The reasoning and alignment scenarios study whether a strong model can learn
from weak critique on a reasoning or alignment task and can improve performance on the in-domain validation set.

\begin{figure*}[t]
    \centering
    \includegraphics[
        width=\textwidth,
        height=0.5\textheight,
        keepaspectratio
    ]{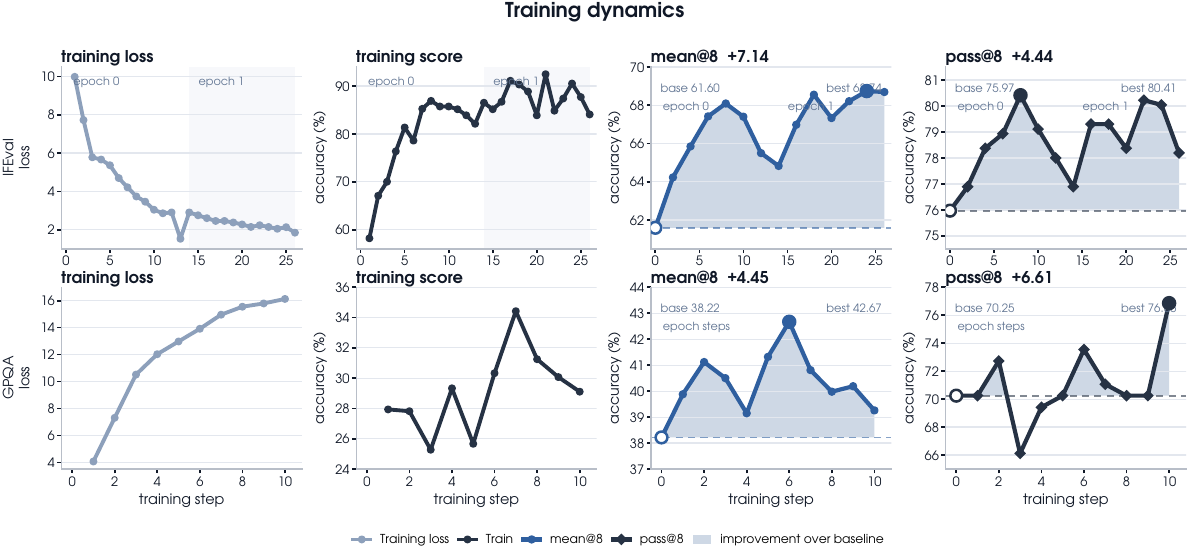}
    \caption{Training dynamics of \ours in alignment and reasoning scenarios. The left column shows results in the alignment scenario, while the right column shows results in the reasoning scenario. For each scenario, we report the training loss, training score, and validation performance over global training steps. The highlighted regions in Validation indicate the improvement over initial validation on the test set before training.}
    \label{fig:OPCD_training}
\end{figure*}

\paragraph{Experimental Setup.}
In the reasoning scenario, we use Qwen3-4B-base~\cite{yang2025qwen3} as the strong model and Qwen3-1.7B-base~\cite{yang2025qwen3} as the weak model. In the alignment scenario, we use Phi-4-14B~\cite{abdin2024phi} as the strong model and Phi-4-mini-instruct~\cite{abdin2024phi} as the weak model. To follow the weak to strong generalization \cite{burns2023weak} setting, we supervise fine-tune (SFT)~\cite{radford2018improving} the weak model in the alignment scenario and use it to provide critiques in the \ours training.

\paragraph{Training and Evaluation Data.}
For reasoning, we use the GPQA main split~\cite{rein2023gpqa} excluding the GPQA-Diamond subset as the source corpus. 
This produces a non-Diamond GPQA subset with 250 multiple-choice science questions. We split this subset into two disjoint parts: 129 questions are used for training, and the remaining 121 questions are held out for in-domain validation. The training split is used to construct weak supervision from Qwen3-1.7B-Base, which is then used to train Qwen3-4B-Base under our weak-to-strong distillation framework. For alignment, we use the Magpie alignment corpus~\cite{xu2024magpiealignmentdatasynthesis} which is IFEval-style ~\cite{zhou2023instruction} dataset as the source training corpus and sample 5,000 instruction-response examples. We randomly split these examples into two disjoint subsets of equal size. The first 2,500 examples are used to fine-tune Phi-4-mini-instruct into a weak alignment supervisor, and the remaining 2,500 examples are used to train Phi-4-14B with critiques generated by this weak model. We use the test split of IFEval~\cite{zhou2023instruction} as the in-domain validation set during training.

\paragraph{Hyperparameters.}
For the alignment experiment, the weak critic is initialized from a Phi-4-mini-instruct checkpoint supervised fine-tuned on the SFT training split with learning rate $1\times 10^{-6}$, and we use the step-40 checkpoint for later \ours training. For both alignment and reasoning experiments, the rollout number in on-policy answer generation stage is $n=8$, the training batch size is 128, the PPO mini-batch size is 32, and the PPO micro-batch size per GPU is 1. The maximum prompt length is 4096. The maximum response length is 8192 for alignment and 6000 for reasoning. We optimize the actor with AdamW using weight decay 0.01 and no learning-rate warmup. The actor learning rate is $1\times 10^{-6}$ for alignment experiment and $5\times 10^{-7}$ for reasoning experiment. At weak critique generation and refinement stage, we sample initial answer rollouts with temperature 1.0, top-$k=20$, and top-$p=0.95$. We use greedy decoding for critique generation, top-$p=1.0$ and top-$k=-1$. We sample refined answer generation with temperature 0.6, top-$k=20$, and top-$p=0.95$. Validation uses 8 sampled responses per prompt with temperature 0.6, top-$k=20$, and top-$p=0.95$. For the alignment setting, we evaluate the trained strong model on IFEval during training and report mean@8 and pass@8. 
For the reasoning setting, we evaluate on the 121 held-out non-Diamond GPQA problems and report the same sampling-based metrics. Here, mean@8 measures the average accuracy over eight sampled responses, while pass@8 measures whether at least one of the eight responses is correct.

\begin{table}[ht]
\centering
\caption{Training dynamics of \ours in the alignment experiment. Accuracy-related metrics, including Train, mean@8, and pass@8, are reported as percentages.}
\label{tab:sdpo-ifeval-metrics}
\small
\renewcommand{\arraystretch}{1.08}
\setlength{\tabcolsep}{3.5pt}
\resizebox{0.9\columnwidth}{!}{
\begin{tabular}{rrrrrr}
\toprule
\textbf{Step} & \textbf{Epoch} & \textbf{Training Score} & \textbf{Train Score} & \textbf{mean@8} & \textbf{pass@8} \\
\midrule
0  & -- & --     & --    & 61.60 & 75.97 \\
1  & 0  & 9.9845 & 58.20 & --    & --    \\
2  & 0  & 7.7260 & 67.09 & 64.23 & 76.89 \\
3  & 0  & 5.7806 & 70.02 & --    & --    \\
4  & 0  & 5.6632 & 76.37 & 65.85 & 78.37 \\
5  & 0  & 5.3658 & 81.35 & --    & --    \\
6  & 0  & 4.6986 & 78.61 & 67.42 & 78.93 \\
7  & 0  & 4.2116 & 85.25 & --    & --    \\
8  & 0  & 3.7364 & 86.91 & 68.09 & \textbf{80.41} \\
9  & 0  & 3.4646 & 85.74 & --    & --    \\
10 & 0  & 3.0429 & 85.74 & 67.40 & 79.11 \\
11 & 0  & 2.8580 & 85.16 & --    & --    \\
12 & 0  & 2.9025 & 83.89 & 65.50 & 78.00 \\
13 & 0  & 1.5260 & 82.12 & --    & --    \\
\midrule
14 & 1  & 2.9028 & 86.52 & 64.83 & 76.89 \\
15 & 1  & 2.7527 & 85.16 & --    & --    \\
16 & 1  & 2.6034 & 86.72 & 66.98 & 79.30 \\
17 & 1  & 2.4570 & 91.11 & --    & --    \\
18 & 1  & 2.4585 & 90.33 & 68.55 & 79.30 \\
19 & 1  & 2.3776 & 88.87 & --    & --    \\
20 & 1  & 2.2747 & 83.89 & 67.33 & 78.37 \\
21 & 1  & 2.1447 & 92.48 & --    & --    \\
22 & 1  & 2.2345 & 84.86 & 68.21 & 80.22 \\
23 & 1  & 2.1421 & 87.40 & --    & --    \\
24 & 1  & 2.0481 & 90.53 & \textbf{68.74} & 80.04 \\
25 & 1  & 2.1318 & 87.70 & --    & --    \\
26 & 1  & 1.8460 & 84.08 & 68.69 & 78.19 \\
\bottomrule
\end{tabular}
}
\end{table}

\paragraph{\ours Improves Training-Time Performance.}
As shown in Table~\ref{tab:sdpo-ifeval-metrics} and Table~\ref{tab:sdpo-gpqa-conf06-metrics}, \ours improves strong-model performance in both alignment and reasoning training. 
On IFEval, the initial model achieves 61.60 mean@8 and 75.97 pass@8, while \ours improves mean@8 to 68.74 and pass@8 to 80.41 during training. 
On GPQA, the initial model achieves 38.22 mean@8 and 70.25 pass@8, while \ours improves mean@8 to 42.67 and pass@8 to 76.86. 
These results show that weak-model critique can provide useful training-time supervision for the strong model across both instruction-following alignment and scientific reasoning tasks. 
This observation is also consistent with our preliminary experiments, where weak-model critiques improve strong-model performance without updating model weights.

\paragraph{Reasoning and Alignment Exhibit Different Training Dynamics.}
Although \ours improves both settings, the training dynamics are substantially different. In the alignment experiment, the training loss (distillation loss) generally decreases from 9.9845 to around 2, and the Train score rises from 58.20 to over 90 at several steps, indicating relatively stable optimization under weak critique supervision. In contrast, in the reasoning experiment, the policy-gradient loss increases from 4.0672 to 16.1311, and the train score fluctuates between 25.28 and 34.42 rather than increasing smoothly. This suggests that reasoning tasks produce noisier and higher-variance weak-to-strong training signals than alignment tasks. A possible reason is that scientific reasoning requires precise multi-step correctness, so weak critiques may help the strong model find useful reasoning directions but do not always provide high-quality supervision for every rollout. Therefore, compared with alignment training, reasoning training benefits from \ours but remains more sensitive to critique quality.

\begin{table}[htbp]
\centering
\caption{Training dynamics of \ours in the reasoning experiment. Accuracy-related metrics, including Train, mean@8, and pass@8, are reported as percentages.}
\label{tab:sdpo-gpqa-conf06-metrics}
\small
\renewcommand{\arraystretch}{1.08}
\setlength{\tabcolsep}{3.5pt}
\resizebox{0.8\columnwidth}{!}{
\begin{tabular}{rrrrrr}
\toprule
\textbf{Epoch} & \textbf{Training Loss} & \textbf{Train Score} & \textbf{mean@8} & \textbf{pass@8} \\
\midrule
 -- & --     & --    & 38.22 & 70.25 \\
 0  & 4.0672 & 27.94 & 39.88 & 70.25 \\
 1  & 7.3085 & 27.82 & 41.12 & 72.73 \\
 2  & 10.5132 & 25.28 & 40.50 & 66.12 \\
 3  & 12.0304 & 29.33 & 39.15 & 69.42 \\
 4  & 12.9736 & 25.67 & 41.32 & 70.25 \\
 5  & 13.9145 & 30.33 & \textbf{42.67} & 73.55 \\
 6  & 14.9623 & 34.42 & 40.81 & 71.07 \\
 7  & 15.5532 & 31.25 & 39.98 & 70.25 \\
 8  & 15.7956 & 30.07 & 40.19 & 70.25 \\
 9  & 16.1311 & 29.11 & 39.26 & \textbf{76.86} \\
\bottomrule
\end{tabular}
}
\end{table}

\section{Conclusion}

We introduce weak-critic strong oversight, a scalable oversight setting where a weak model guides a stronger model through critiques rather than labels, preferences, or final judgments. This reduces the burden on the weak supervisor, since it only needs to provide a useful and non-misleading revision direction instead of solving or judging the full task. Our inference-time experiments show that weak critiques can improve frozen strong models on both reasoning and alignment benchmarks, and that critique quality is central to reliable improvement. Motivated by this finding, we proposed On-Policy Critique Distillation (\ours), which filters high-quality weak critiques and distills critic-guided behavior into the strong model using adaptive self-teacher signals. Training-time results on IFEval and GPQA show that \ours can improve strong-model behavior over training steps, demonstrating that weak critiques can provide effective supervision even when direct weak labels or judgments are unreliable. These findings suggest that critique-based weak supervision is a promising path for weak-to-strong generalization and scalable oversight. Future work can further improve critique-quality estimation, study broader task domains, and combine weak critiques with other oversight mechanisms such as debate, process supervision, or verifier-assisted training.

\bibliography{example_paper}
\bibliographystyle{icml2026}

\newpage
\appendix
\onecolumn


\end{document}